\documentclass[pdflatex,sn-mathphys-num]{sn-jnl}

\usepackage{graphicx}
\usepackage{multirow}
\usepackage{amsmath,amssymb,amsfonts}
\usepackage{amsthm}
\usepackage{mathrsfs}
\usepackage[title]{appendix}
\usepackage[table]{xcolor}  
\usepackage{textcomp}
\usepackage{manyfoot}
\usepackage{booktabs}
\usepackage{algorithm}
\usepackage{algorithmicx}
\usepackage{algpseudocode}
\usepackage{listings}
\usepackage{array}
\usepackage{pdflscape}
\usepackage{rotating}
\usepackage{tikz}
\usetikzlibrary{shapes, arrows, positioning}

\newcolumntype{P}[1]{>{\raggedright\arraybackslash}p{#1}}

\theoremstyle{thmstyleone}%
\theoremstyle{thmstyletwo}%

\theoremstyle{thmstylethree}%

\begin{document}

\title[Article Title]{Reversing the Paradigm: Building AI-First Systems with Human Guidance}


\author*[1]{\fnm{Cosimo} \sur{Spera}}\email{cosimo@minervacq.com}

\author[1]{\fnm{Garima} \sur{Agrawal}}\email{garima@minervacq.com}
\equalcont{These authors contributed equally to this work.}

\affil*[1]{\orgname{Minerva CQ}, \country{USA}}



\abstract{The relationship between humans and artificial intelligence is no longer a concept from science fiction—it’s a growing reality reshaping how we live and work. Today, AI isn’t just confined to research labs or tech giants’ headquarters; it’s embedded in everyday experiences, quietly powering customer service chats, personalizing travel recommendations, assisting doctors with diagnoses, and supporting educators in classrooms.

What makes this moment in AI’s evolution particularly compelling is its increasing collaborative nature. Rather than replacing humans, AI is augmenting our capabilities—streamlining routine tasks, enhancing decision-making with data-driven insights, and fueling creativity in fields like design, music, and writing. The future of work is shifting toward AI agents taking the lead on tasks, with humans stepping in as supervisors, strategists, and ethical stewards—flipping the traditional model on its head. Instead of humans driving every action with AI as a support tool, intelligent agents will increasingly operate autonomously within defined parameters, handling everything from scheduling and customer interactions to complex decision-making workflows. Humans will guide, monitor, and fine-tune these agents to ensure alignment with organizational goals, values, and context. This shift promises significant gains in efficiency and scalability but also demands a new mindset—one that prioritizes meaningful collaboration between autonomous AI systems and thoughtful human guidance.

As AI agents assume more autonomous roles, the potential benefits are substantial: increased productivity, accelerated decision-making, cost savings, and the ability to scale operations in unprecedented ways. However, these benefits come with notable risks—including reduced human control, algorithmic bias, security vulnerabilities, and a widening skills gap. To navigate this transition successfully, organizations and societies must rethink roles, invest in upskilling, embed ethical frameworks into AI development, and prioritize transparency and accountability.

This paper explores the technological and organizational shifts required to responsibly transition to AI-first systems—where autonomy is balanced with human values, guidance, and strategic intent.}







\maketitle

\noindent\textbf{Paper Organization}

This paper begins with an overview of the current AI landscape, followed by a discussion of the prevailing paradigm: “Humans in the driver’s seat, with AI in a supporting role.” The third section presents the case for reversing this paradigm—advocating for AI-first systems where humans provide guidance rather than direct control. The fourth and fifth sections explore the design principles for such systems and examine associated risks. The sixth section highlights current AI-first use cases, while the seventh outlines a timeline for transitioning toward this model. The paper concludes with reflections on designing responsible, scalable AI-first systems guided by human values.

\section{Introduction}\label{sec1}

The emergence of generative AI and autonomous systems marks a significant technological inflection point, with far-reaching implications across domains. Generative AI models—such as large language models and generative adversarial networks—demonstrate unprecedented capabilities in producing coherent text, realistic images, and functional code, effectively automating aspects of creativity and problem-solving. Concurrently, autonomous systems—ranging from robotics and drones to self-driving vehicles—are becoming increasingly adept at perceiving their environments, making decisions, and executing actions without direct human involvement. These advances promise increased efficiency and innovation but also raise challenges related to safety, accountability, and the socio-economic consequences of automation.

The current AI paradigm primarily positions artificial intelligence as a tool or assistant to human operators. In this model, AI augments human capabilities—supporting decision-making, automating routine tasks, and extracting insights from data—while humans retain primary control and responsibility. This \textit{human-in-the-loop} approach emphasizes interpretability, transparency, and ethical safeguards, ensuring that AI operates within boundaries defined by human judgment. The assistant paradigm has enabled productivity gains across sectors such as healthcare, finance, education, and customer service, while mitigating risks associated with full autonomy.

An emerging shift in AI research and deployment envisions a reversal of this traditional human-centric model. In this evolving paradigm, AI systems act as the primary operators, with humans stepping into guiding, supervisory, and ethical roles. This inversion recognizes the increasing capability of autonomous AI systems to manage complex, high-frequency decisions with speed, scalability, and consistency. In such systems, humans no longer drive every action but instead provide contextual judgment, strategic input, and system-level intervention when necessary. This transformation in roles has profound implications for how we design, govern, and interact with AI systems—demanding a new framework for trust calibration, accountability, and collaboration in AI-first environments.

We define \textit{AI-first systems} as socio-technical architectures in which AI agents serve as the primary operational entities—autonomously performing tasks, making decisions, and coordinating workflows—while humans adopt supervisory, strategic, or ethical guidance roles. Unlike traditional human-in-the-loop models, where AI augments human decision-making, AI-first systems invert this dynamic by embedding autonomy into the core execution layer, with humans guiding and refining system-level behavior as needed. This design shift enables greater scalability, responsiveness, and efficiency in domains where algorithmic decision-making can outperform human-led processes.

\section{The Current Paradigm: Humans in the Driver’s Seat}\label{sec2}

AI technologies have become deeply integrated into everyday digital experiences through use cases such as copilots, chatbots, and recommendation engines. Copilot systems—exemplified by tools like GitHub Copilot or AI writing assistants—augment human productivity by generating context-aware suggestions in real time, streamlining tasks in software development, writing, and data analysis. Chatbots, powered by natural language processing, serve as front-line interfaces in customer service, healthcare triage, and education, offering scalable, round-the-clock engagement while reducing the burden on human agents. Recommendation engines, widely deployed in e-commerce, entertainment, and social media platforms, personalize user experiences by predicting preferences based on behavioral data, thereby increasing engagement and optimizing content delivery. Collectively, these applications illustrate how AI functions as an assistive layer—enhancing efficiency, personalization, and accessibility across a range of domains.

Positioning AI as an assistant, with humans in the lead decision-making role, offers several practical advantages—particularly in terms of control, trust, and legal clarity. First, it ensures that human operators maintain ultimate authority over key decisions, with AI confined to supporting functions within well-defined boundaries. Second, trust is reinforced when AI behavior remains interpretable, predictable, and subject to human validation, allowing for real-time corrections and calibration. This structure also simplifies accountability: when humans are explicitly in the driver’s seat, it becomes easier to assign responsibility, ensure regulatory compliance, and enforce ethical standards. These characteristics have made the current paradigm well-suited to environments where transparency, caution, and institutional trust are paramount.

However, this model is not without limitations—particularly as AI capabilities continue to scale. Human guidance in high-stakes or ambiguous scenarios can become a bottleneck, constraining the scalability and responsiveness of AI systems. Latency may increase when decisions are deferred to human input, especially in time-sensitive contexts. Supervisors may experience cognitive overload when required to monitor multiple AI agents or interpret complex outputs without adequate abstraction and summarization tools. Furthermore, the continuous operation of advanced AI models raises concerns about energy consumption and long-term sustainability. Finally, deploying and maintaining such hybrid systems requires significant investment in infrastructure, personnel training, and human-in-the-loop frameworks—factors that may limit widespread adoption across industries.

\section{The Case for Reversal: Why and When AI Should Lead}\label{sec3}

AI should be positioned as the lead operator in socio-technical systems when the capabilities of artificial agents surpass human limitations in processing speed, sensory integration, and the ability to sustain performance over extended periods without fatigue or bias. This threshold is increasingly being met due to rapid advances in foundational AI technologies. Large Language Models (LLMs), such as GPT-4 and its successors, now demonstrate sophisticated language comprehension, contextual reasoning, and generative capabilities—enabling them to perform complex tasks such as drafting legal documents, responding to technical inquiries, or managing customer interactions with minimal human intervention. Planning agents, often built upon reinforcement learning or neuro-symbolic architectures, can sequence actions, optimize for long-term objectives, and adapt dynamically to changing constraints. Simultaneously, breakthroughs in multi-modal perception empower AI agents to interpret and integrate data from diverse sources—text, speech, vision, and structured inputs—enabling more situationally aware operation. Complementing these capabilities are memory-augmented models that maintain persistent knowledge over time, supporting continuity across tasks, interactions, and decision cycles. Together, these innovations enable AI systems to operate autonomously in environments where information is vast, response times are critical, and consistency is paramount.

From an economic perspective, the shift toward AI-led operations promises significant efficiency gains and cost reductions. Automating routine, high-volume, and knowledge-intensive tasks allows organizations to scale without linear increases in labor costs, reduce human error, and reallocate skilled professionals to supervisory, creative, or strategic functions. In sectors such as finance, logistics, healthcare, and e-commerce, this transformation supports faster service delivery, improved resource utilization, and the creation of new revenue models driven by real-time, personalized AI services. Moreover, once deployed, AI systems offer non-linear returns on investment, as their marginal operating costs are low relative to the fixed costs of development and integration. Thus, from both a technical and economic standpoint, AI should assume operational leadership in domains where tasks are suited to algorithmic optimization, adaptive learning, and scalable deployment.

One of the most compelling advantages of AI-first systems lies in their ability to automate complex workflows that historically required highly skilled human labor. LLMs can now interpret nuanced instructions and generate domain-specific content, while planning agents can coordinate multi-step processes autonomously. These capabilities allow AI to manage intricate workflows across areas such as legal analysis, medical triage, supply chain optimization, and software engineering—reducing dependency on costly human expertise, minimizing delays, and eliminating inefficiencies. Multi-modal AI further extends this potential by integrating diverse data streams—text, images, audio, and structured data—within a unified automated pipeline, thereby eliminating handoffs between specialized teams. Memory-enabled agents retain institutional knowledge across interactions, reducing the need for repeated onboarding and training. This reconfiguration of traditional workflows drives non-linear scalability and long-term savings while maintaining, or even enhancing, quality and responsiveness.

Strategically, AI-led systems confer a decisive advantage in data-intensive and time-sensitive domains. In finance, AI agents can autonomously analyze market signals, execute trades, detect fraud, and generate real-time risk assessments at a speed and volume beyond human capacity. In logistics, AI can orchestrate global supply chains—dynamically optimizing routes, inventory levels, and forecasts while adapting to disruptions with minimal human involvement. Customer service operations increasingly rely on AI agents capable of resolving high volumes of queries with contextual accuracy and personalized engagement. In intelligence and national security, AI can synthesize multimodal data, identify patterns, and coordinate cyber-defense and surveillance operations across distributed assets. These benefits stem from the ability to orchestrate AI systems that operate continuously, learn from interaction, and respond coherently in complex, dynamic environments. Centralizing operational and cognitive capabilities in AI agents enables organizations to reduce latency, increase adaptability, and outperform competitors—offering a structural advantage that human-led workflows alone cannot match.

While the case for AI-first systems is strong, it is essential to acknowledge the accompanying risks. The misuse of advanced AI—particularly in generating deceptive content, automating cyberattacks, or perpetuating bias at scale—poses real societal challenges. As such, the transition to AI-led systems must be accompanied by strong safeguards, continuous human guidance, and governance mechanisms that ensure alignment with ethical and legal standards.

\section{Designing Human-Supported AI Frameworks}\label{sec4}

Human supervisors play a critical role in the guidance and augmentation of AI systems, particularly in domains where ambiguity, ethical sensitivity, and exception handling are common. Their ability to intervene during unexpected conditions ensures robust system performance beyond the capabilities of pre-programmed responses. Moreover, human ethical supervision is indispensable for evaluating the broader societal implications of automated decisions—especially in high-stakes domains such as healthcare, criminal justice, or autonomous transportation. Supervisors also provide critical support in reasoning about edge cases, where data may be sparse or non-representative, enabling context-aware responses to atypical scenarios. Furthermore, subjective human judgment—grounded in empathy, lived experience, and cultural awareness—remains essential in contexts where algorithmic reasoning alone is insufficient. These human contributions are integral to ensuring the safety, fairness, and accountability of complex AI-first systems.

Governance and auditability are foundational pillars for deploying AI systems responsibly. These pillars ensure transparency, accountability, and alignment with human values throughout the system lifecycle. Transparency enables stakeholders to inspect and understand how decisions are made—requiring not just visibility into system architectures and data pipelines, but also clear documentation of system goals, design assumptions, and limitations. Explainability complements this by making AI outputs interpretable to both technical and non-technical users, thereby fostering trust and facilitating regulatory compliance. Continuous feedback loops support auditability by allowing real-world monitoring, adaptation, and refinement of system performance. These loops enable retrospective evaluations and prospective improvements, forming a dynamic governance structure.

While Human-in-the-Loop (HITL) is traditionally associated with human-led systems, in AI-first frameworks it serves as a supervisory mechanism—where humans intervene selectively to guide autonomous AI agents, rather than directing every decision. HITL principles remain critical in embedding human supervision at key decision points, ensuring that AI systems remain accountable and aligned with ethical, legal, and institutional expectations.

Interface design plays a pivotal role in enabling effective human supervision of AI systems. Rather than focusing solely on execution, interfaces must support supervisory control and situational awareness. User interfaces (UIs) and application programming interfaces (APIs) should be designed to present information clearly, contextually, and hierarchically—enabling supervisors to detect anomalies, understand system states, and make informed decisions. Alerting mechanisms must balance precision and relevance, avoiding both under reporting and user fatigue. Well-calibrated intervention thresholds and structured escalation policies ensure that the right level of attention is directed to the right issues at the right time. These interface design elements foster a collaborative environment where human supervisors act as informed stewards, enabling safe and scalable deployment of autonomous AI systems.

A practical example of this approach is demonstrated by Minerva CQ’s Human-in-the-Loop solution for contact center operations \cite{agrawal2024beyondrag}. Minerva CQ combines real-time conversational AI with human expertise to enhance customer-agent interactions. The system analyzes live calls, providing agents with timely prompts, relevant knowledge, and decision support. While AI accelerates resolution and augments decisions, humans retain the authority to intervene, override, or refine AI suggestions. This synergy reduces cognitive load, increases customer satisfaction, and enables continuous model improvement through feedback loops—an essential feature in high-volume, high-variability environments like customer support. While Minerva CQ operates as a HITL solution today, it exemplifies how AI-first systems can embed human guidance without compromising autonomy or efficiency.

Table~\ref{tab:kpis} highlights example KPIs reported by contact centers using Minerva CQ’s agent-assist solution. These results underscore the practical value of HITL systems in improving both agent performance and customer experience.

\begin{table}[h]
\centering
\caption{Contact Center HITL Solution – KPI Improvements}
\label{tab:kpis}
\begin{tabular}{|l|l|}
\hline
\textbf{KPI} & \textbf{Improvement} \\
\hline
Average Handle Time (AHT) & Reduction of 20–40\% \\
First Call Resolution (FCR) & Increase by 10–20\% \\
Customer Satisfaction (CSAT) & Increase by 10–15\% \\
Agent Assist Accuracy & Improves from 85–90\% to 95\%+ via feedback \\
Training Ramp Time & Reduction of 30–50\% \\
Call Deflection & Increase of 10–15\% through improved self-service \\
Agent Turnover Rate & Reduction of 10–25\% \\
\hline
\end{tabular}
\end{table}

These KPI improvements vary by industry and deployment scale but represent consistent gains enabled by effective human-supervised AI systems. Such frameworks exemplify how thoughtful integration of human supervision can enhance performance, maintain accountability, and realize the full potential of AI-first architectures.

\section{Risks and Mitigations}\label{sec5}

The deployment of advanced AI systems within critical decision-making contexts introduces a spectrum of risks that necessitate comprehensive mitigation through systemic, technical, and regulatory interventions. Foremost among these is the potential erosion of effective human supervision, particularly in high-autonomy environments where AI agents may exhibit opaque or unpredictable behavior. To address this, it is essential to implement supervisory mechanisms such as traceable decision logs, formal verification of system behaviors, and interactive simulations that support human understanding and timely intervention. These tools enable human supervisors to examine, test, and approve AI-driven actions either prior to deployment or during real-time operations.

A second major concern involves the amplification of bias and harm—often arising from machine learning models trained on incomplete, imbalanced, or historically biased datasets. Such systems risk entrenching existing social inequities or introducing novel forms of discrimination. Mitigation strategies must therefore include the use of diverse and representative training data, robust fairness auditing tools, and continuous monitoring of system outputs for disparate impacts. Additionally, fallback mechanisms—such as deferring to human judgment or invoking rule-based overrides—should be integrated when predefined fairness thresholds are violated.

The displacement of human labor is another pressing risk, particularly as AI systems increasingly take on tasks that were traditionally performed by skilled cognitive workers. Addressing this challenge requires not only minimizing job displacement but also enabling meaningful workforce transition through targeted reskilling programs. Organizational restructuring should prioritize human–AI collaboration by redefining human roles around system supervision, design, and strategic guidance, rather than replacement.

Finally, the issue of legal accountability presents a complex challenge, given the distributed responsibility among developers, human supervisors, and autonomous systems themselves. Resolving this requires regulatory innovation, including the development of new liability frameworks capable of apportioning accountability in cases of system failure or harm. This may involve revisiting legal constructs such as product liability, establishing auditable certification schemes for AI systems, and enforcing transparency requirements that ensure decision traceability across increasingly intricate socio-technical infrastructures.

\begin{figure}[htbp]
\centering
\begin{tikzpicture}[
  node distance=0.75cm,
  every node/.style={align=center, font=\footnotesize},
  layer/.style={draw, minimum height=1.2cm, text width=10.8cm, 
                inner sep=6pt, font=\footnotesize, line width=1pt},
  arrow/.style={->, >=latex, line width=1.8pt},
  governance/.style={layer, fill=red!12, draw=red!80},
  supervision/.style={layer, fill=orange!10, draw=orange!70},
  ai/.style={layer, fill=blue!15, draw=blue!80},
  task/.style={layer, fill=green!15, draw=green!80}
]

\node[governance] (gov) {
  \textbf{Governance, Feedback, and Ethical Oversight Layer} \\
  \textit{Risk monitoring, fallback mechanisms, audit, certification, policy enforcement}
};
\node[supervision, below=of gov] (supervision) {
  \textbf{Human Supervision \& Interface Layer} \\
  \textit{Monitoring, guiding, strategic intervention, UIs, alerts, control panels, adaptive dashboards}
};
\node[ai, below=of supervision] (ai) {
  \textbf{AI-First System Layer} \\
  \textit{Planning agents, LLMs, perception modules, memory-based reasoning, decision making}
};
\node[task, below=of ai] (tasks) {
  \textbf{Autonomous Execution of Domain-Specific Tasks} \\
  \textit{Customer support, trading, diagnostics, data processing, content generation}
};

\draw[arrow, red!70] (gov) -- (supervision);
\draw[arrow, orange!70] (supervision) -- (ai);
\draw[arrow, blue!70] (ai) -- (tasks);

\draw[arrow, green!60]
  (tasks.east) .. controls +(2.2,0) and +(2.2,0) ..
  (gov.east);

\node[font=\bfseries\footnotesize, text=green!50!black, rotate=90, right=2.4cm of ai.east] 
  {Feedback Loop};

\end{tikzpicture}
\caption{Layered architecture of AI-first systems with governance and human oversight}
\label{fig:ai-layered-architecture}
\end{figure}
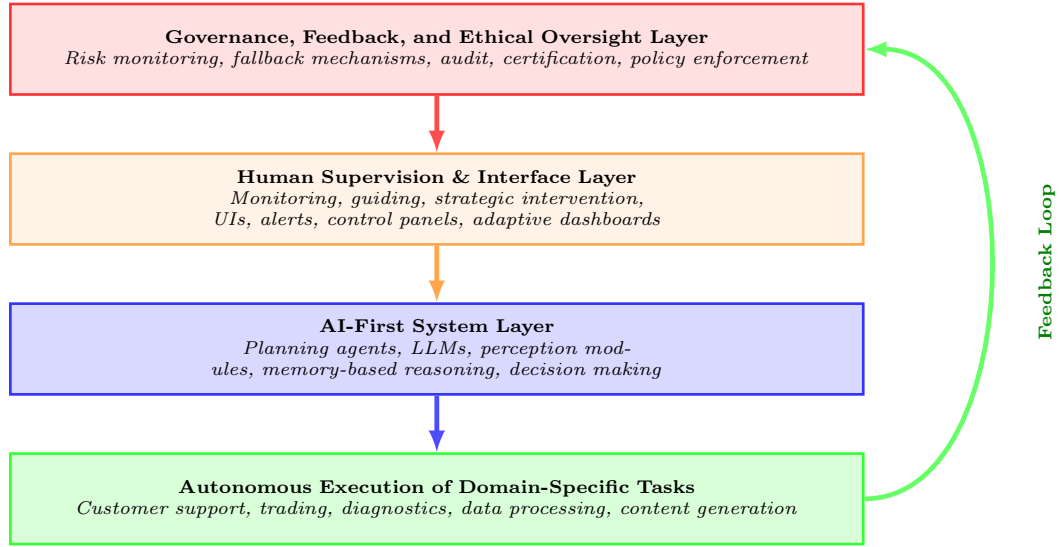

\section{Case Studies and Prototypes}\label{sec6}

AI-first approaches are increasingly being deployed in high-impact operational domains, offering both efficiency gains and new challenges in supervision and accountability. Autonomous trading systems, for example, operate at millisecond timescales in global financial markets, using machine learning models to detect patterns, execute trades, and adapt strategies in real time without direct human involvement. While these systems have improved market liquidity and responsiveness, they have also introduced risks such as flash crashes and systemic instabilities, underscoring the need for built-in monitoring and fail-safe mechanisms.

In the domain of self-healing IT infrastructure, AI-driven platforms autonomously detect anomalies, predict failures, and initiate corrective actions—such as re-routing traffic or restarting services—often before human supervisors are even alerted. This has led to significant improvements in uptime and operational resilience but also raises concerns about over-reliance on opaque decision-making in mission-critical environments.

Similarly, AI-led customer service systems now handle large volumes of interactions across sectors such as banking, healthcare, and telecommunications, using natural language processing to interpret queries, resolve issues, and escalate to human agents only when necessary. These systems exemplify human-in-the-loop implementations within AI-first architectures, where AI autonomously manages routine interactions and humans intervene in complex or sensitive cases.

From these examples, several lessons emerge: fully autonomous systems can deliver exceptional performance, but they must be complemented by robust monitoring, explainability, and escalation mechanisms. Moreover, AI-first does not mean AI-only—effective deployment often relies on hybrid architectures in which humans retain meaningful supervisory roles. Looking ahead, responsible AI integration will require adaptive governance frameworks, redefined workforce responsibilities, and continuous alignment of AI systems with evolving ethical, social, and legal norms. The architecture in Figure~\ref{fig:ai-layered-architecture} exemplifies how AI-first systems can operate autonomously while remaining embedded within layers of human supervision and governance, ensuring ethical, adaptable, and task-specialized performance.

\section{Strategic Roadmap for Responsible Integration of AI-First Systems}\label{sec7}

The integration of AI-first systems into complex organizational workflows demands a phased strategic roadmap that aligns technological advancement with responsible governance. A staged approach—structured across short-, medium-, and long-term horizons—enables organizations to manage risk while progressively transforming operational paradigms.

\subsection*{Short Term (1–2 Years)}
In the short term (1–2 years), efforts should focus on identifying high-impact, low-risk application areas where AI can deliver immediate value with minimal ethical or safety implications. Typical domains include internal process optimization, IT operations, and customer service triage. During this phase, emphasis should be placed on developing functional prototypes embedded with human guardrails—such as supervision mechanisms, intervention pathways, and interpretable decision-making frameworks. These implementations not only validate technical viability but also serve to build institutional capacity, foster cross-functional collaboration, and cultivate trust among end users.

\subsection*{Medium Term (3–5 Years)}
The medium-term phase (3–5 years) emphasizes institutionalizing lessons from early pilots through workforce transformation and governance refinement. This includes designing role-evolution strategies and training programs that empower employees to supervise, audit, and co-work with AI systems. Priorities include the deployment of real-time monitoring tools, feedback loops for continuous learning, and adaptive interfaces that enhance human comprehension and engagement. As AI systems take on more complex and interdependent responsibilities, system transparency and trust become critical enablers of scalable adoption.

\subsection*{Long Term (5–10 Years)}
In the long term (5–10 years), the strategic objective shifts toward enabling policy innovation and achieving constrained autonomy in clearly defined domains such as logistics, predictive maintenance, or diagnostic workflows. As seen in earlier case studies—such as autonomous trading systems and self-healing IT infrastructures—long-term integration requires not only advanced capabilities but also robust cross-domain coordination.

To support this, organizations should establish federated supervision frameworks, wherein local teams retain oversight of domain-specific AI agents while centralized governance ensures enterprise-wide alignment with regulatory and ethical norms. This distributed model of human supervision promotes consistency, scalability, and responsiveness across multiple AI-first deployments.

This roadmap outlines a scalable and adaptive strategy for transitioning from pilot deployments to sustainable, AI-first infrastructures that remain aligned with human values and institutional accountability.

\begin{table}[ht]
\centering
\caption{Strategic roadmap for responsible integration of AI-first systems with impact mapping}
\label{tab:roadmap-impact}
\rowcolors{2}{gray!10}{white} 
\begin{tabular}{
|>{\columncolor{gray!20}}p{1.5cm}|
>{\columncolor{gray!20}}p{2cm}|
>{\columncolor{gray!20}}p{3.2cm}|
>{\columncolor{gray!20}}p{2cm}|
>{\columncolor{gray!20}}p{3cm}|}
\hline
\textbf{Time horizon} & \textbf{Strategic focus} & \textbf{Key actions} & \textbf{Impact area} & \textbf{Expected outcomes} \\
\hline
Short term (1--2 yrs) & 
Cautious deployment in low-risk, high-value domains & 
Build AI-first prototypes with human supervision; ensure interpretability and trust & 
Process optimization, user trust & 
Validated use-cases; Increased organizational AI readiness \\
\hline
Medium term (3--5 yrs) & 
Organizational evolution and trust-building & 
Redesign roles; scale training; deploy adaptive interfaces; real-time monitoring & 
Workforce, governance & 
Higher adoption; Workforce AI fluency; Auditability \\
\hline
Long term (5--10 yrs) & 
Policy adaptation and constrained autonomy & 
Federated oversight; coordination of multi-agent AI; regulatory alignment & 
Cross-domain AI governance & 
Sustainable AI deployment; Ethical compliance; System-level coordination \\
\hline
\end{tabular}
\end{table}

\section{Conclusion and Next Steps}

The transition to AI-first systems represents a profound paradigm shift—one that goes beyond technological innovation to fundamentally redefine the relationship between humans and AI systems. The objective is not to pursue AI supremacy, but to cultivate a symbiotic partnership in which automation augments human insight, and human supervision ensures the ethical and accountable operation of autonomous agents. This vision demands a multi-disciplinary, cross-sectoral commitment to advancing foundational research, accelerating responsible innovation, and constructing adaptive governance architectures resilient to the evolving demands of AI integration.

To shape a future in which AI systems advance collective human interests, design and deployment decisions must be guided not solely by performance metrics but by long-term values such as resilience, fairness, and shared agency. The success of this transition hinges on our ability to embed human values into technical systems from the outset—ensuring future architectures are not only intelligent, but also just, accountable, and societally aligned.

To operationalize this vision, we recommend the following strategic actions:

\begin{enumerate}
    \item \textbf{Establish Cross-Functional AI Governance Councils:}  
    Create interdisciplinary teams comprising technologists, ethicists, legal experts, and domain leaders to guide AI deployment standards, risk management strategies, and policy formation.
    
    \item \textbf{Invest in Human-Centered Design and Supervision Mechanisms:}  
    Prioritize systems that enable transparency, interpretability, and human-in-the-loop (HITL) capabilities across varying levels of autonomy.
    
    \item \textbf{Scale Workforce Development and Reskilling Programs:}  
    Launch targeted educational initiatives to prepare current and future employees for hybrid roles in AI-augmented environments, focusing on AI supervision, validation, and ethical design.
    
    \item \textbf{Promote Open Research and Shared Evaluation Standards:}  
    Support open science and the development of benchmarking frameworks that assess not only technical performance but also societal robustness and ethical alignment.
    
    \item \textbf{Develop and Pilot Sector-Specific AI Certification Schemes:}  
    Collaborate with regulatory bodies and industry consortia to prototype audit-ready certification models that build public trust and ensure operational safety and compliance.
    
    \item \textbf{Design for Scalable, Federated Supervision Architectures:}  
    Anticipate the complexity of multi-agent systems by developing governance frameworks that enable distributed human supervision and coordinated control across AI-first deployments.
\end{enumerate}

These next steps provide a strategic foundation for realizing AI-first systems that are not only autonomous and efficient but also human-aligned, trustworthy, and adaptable to the evolving needs of society.

\bmhead{Acknowledgements}
We would like to express our sincere gratitude to Prof. Franco P. Preparata and Prof. Hal Varian for their support and valuable comments, which greatly enriched the quality of this work.

\nocite{*}

\bibliography{sn-bibliography}


\begin{thebibliography}{35}
\ifx \bisbn   \undefined \def \bisbn  #1{ISBN #1}\fi
\ifx \binits  \undefined \def \binits#1{#1}\fi
\ifx \bauthor  \undefined \def \bauthor#1{#1}\fi
\ifx \batitle  \undefined \def \batitle#1{#1}\fi
\ifx \bjtitle  \undefined \def \bjtitle#1{#1}\fi
\ifx \bvolume  \undefined \def \bvolume#1{\textbf{#1}}\fi
\ifx \byear  \undefined \def \byear#1{#1}\fi
\ifx \bissue  \undefined \def \bissue#1{#1}\fi
\ifx \bfpage  \undefined \def \bfpage#1{#1}\fi
\ifx \blpage  \undefined \def \blpage #1{#1}\fi
\ifx \burl  \undefined \def \burl#1{\textsf{#1}}\fi
\ifx \doiurl  \undefined \def \doiurl#1{\url{https://doi.org/#1}}\fi
\ifx \betal  \undefined \def \betal{\textit{et al.}}\fi
\ifx \binstitute  \undefined \def \binstitute#1{#1}\fi
\ifx \binstitutionaled  \undefined \def \binstitutionaled#1{#1}\fi
\ifx \bctitle  \undefined \def \bctitle#1{#1}\fi
\ifx \beditor  \undefined \def \beditor#1{#1}\fi
\ifx \bpublisher  \undefined \def \bpublisher#1{#1}\fi
\ifx \bbtitle  \undefined \def \bbtitle#1{#1}\fi
\ifx \bedition  \undefined \def \bedition#1{#1}\fi
\ifx \bseriesno  \undefined \def \bseriesno#1{#1}\fi
\ifx \blocation  \undefined \def \blocation#1{#1}\fi
\ifx \bsertitle  \undefined \def \bsertitle#1{#1}\fi
\ifx \bsnm \undefined \def \bsnm#1{#1}\fi
\ifx \bsuffix \undefined \def \bsuffix#1{#1}\fi
\ifx \bparticle \undefined \def \bparticle#1{#1}\fi
\ifx \barticle \undefined \def \barticle#1{#1}\fi
\bibcommenthead
\ifx \bconfdate \undefined \def \bconfdate #1{#1}\fi
\ifx \botherref \undefined \def \botherref #1{#1}\fi
\ifx \url \undefined \def \url#1{\textsf{#1}}\fi
\ifx \bchapter \undefined \def \bchapter#1{#1}\fi
\ifx \bbook \undefined \def \bbook#1{#1}\fi
\ifx \bcomment \undefined \def \bcomment#1{#1}\fi
\ifx \oauthor \undefined \def \oauthor#1{#1}\fi
\ifx \citeauthoryear \undefined \def \citeauthoryear#1{#1}\fi
\ifx \endbibitem  \undefined \def \endbibitem {}\fi
\ifx \bconflocation  \undefined \def \bconflocation#1{#1}\fi
\ifx \arxivurl  \undefined \def \arxivurl#1{\textsf{#1}}\fi
\csname PreBibitemsHook\endcsname

\bibitem[\protect\citeauthoryear{Agrawal et~al.}{2024}]{agrawal2024beyondrag}
\begin{botherref}
\oauthor{\bsnm{Agrawal}, \binits{G.}},
\oauthor{\bsnm{Gummuluri}, \binits{S.}},
\oauthor{\bsnm{Spera}, \binits{C.}}:
Beyond-rag: Question identification and answer generation in real-time conversations.
arXiv preprint arXiv:2410.10136
(2024)
\end{botherref}
\endbibitem

\bibitem[\protect\citeauthoryear{Amodei et~al.}{2016}]{amodei2016concrete}
\begin{botherref}
\oauthor{\bsnm{Amodei}, \binits{D.}},
\oauthor{\bsnm{Olah}, \binits{C.}},
\oauthor{\bsnm{Steinhardt}, \binits{J.}},
\oauthor{\bsnm{Christiano}, \binits{P.}},
\oauthor{\bsnm{Schulman}, \binits{J.}},
\oauthor{\bsnm{Man{\'e}}, \binits{D.}}:
Concrete problems in ai safety.
arXiv preprint arXiv:1606.06565
(2016)
\end{botherref}
\endbibitem

\bibitem[\protect\citeauthoryear{LeCun}{2023}]{lecun2023path}
\begin{botherref}
\oauthor{\bsnm{LeCun}, \binits{Y.}}:
A path towards autonomous machine intelligence.
arXiv preprint arXiv:2301.00250
(2023).
Meta AI blog
\end{botherref}
\endbibitem

\bibitem[\protect\citeauthoryear{Sutton}{2019}]{sutton2019bitter}
\begin{botherref}
\oauthor{\bsnm{Sutton}, \binits{R.S.}}:
The Bitter Lesson.
\url{https://www.incompleteideas.net/}.
Accessed: 2025-06-09
(2019)
\end{botherref}
\endbibitem

\bibitem[\protect\citeauthoryear{Marcus and Davis}{2019}]{marcus2019rebooting}
\begin{bbook}
\bauthor{\bsnm{Marcus}, \binits{G.}},
\bauthor{\bsnm{Davis}, \binits{E.}}:
\bbtitle{Rebooting AI: Building Artificial Intelligence We Can Trust}.
\bpublisher{Pantheon}, \blocation{???}
(\byear{2019})
\end{bbook}
\endbibitem

\bibitem[\protect\citeauthoryear{Huang et~al.}{2021}]{huang2021ai}
\begin{botherref}
\oauthor{\bsnm{Huang}, \binits{T.}}, et al.:
Ai system design: Technical frameworks for scaling ai-first products.
Proceedings of the ACM on Human-Computer Interaction
\textbf{5}(CSCW2)
(2021)
\end{botherref}
\endbibitem

\bibitem[\protect\citeauthoryear{Binns et~al.}{2018}]{binns2018reducing}
\begin{bchapter}
\bauthor{\bsnm{Binns}, \binits{R.}},
\bauthor{\bsnm{Veale}, \binits{M.}},
\bauthor{\bsnm{Van~Kleek}, \binits{M.}},
\bauthor{\bsnm{Shadbolt}, \binits{N.}}:
\bctitle{'it's reducing a human being to a percentage': Perceptions of justice in algorithmic decisions}.
In: \bbtitle{CHI '18: Proceedings of the 2018 CHI Conference on Human Factors in Computing Systems}
(\byear{2018})
\end{bchapter}
\endbibitem

\bibitem[\protect\citeauthoryear{Shneiderman}{2020}]{shneiderman2020human}
\begin{barticle}
\bauthor{\bsnm{Shneiderman}, \binits{B.}}:
\batitle{Human-centered artificial intelligence: Reliable, safe \& trustworthy}.
\bjtitle{International Journal of Human–Computer Interaction}
\bvolume{36}(\bissue{6}),
\bfpage{495}--\blpage{504}
(\byear{2020})
\end{barticle}
\endbibitem

\bibitem[\protect\citeauthoryear{Holstein et~al.}{2019}]{holstein2019fairness}
\begin{bchapter}
\bauthor{\bsnm{Holstein}, \binits{K.}},
\bauthor{\bsnm{Wortman~Vaughan}, \binits{J.}},
\bauthor{\bsnm{Daum{\'e}~III}, \binits{H.}},
\bauthor{\bsnm{Dudik}, \binits{M.}},
\bauthor{\bsnm{Wallach}, \binits{H.}}:
\bctitle{Improving fairness in machine learning systems: What do industry practitioners need?}
In: \bbtitle{CHI '19: Proceedings of the 2019 CHI Conference on Human Factors in Computing Systems}
(\byear{2019})
\end{bchapter}
\endbibitem

\bibitem[\protect\citeauthoryear{Amershi et~al.}{2014}]{amershi2014power}
\begin{barticle}
\bauthor{\bsnm{Amershi}, \binits{S.}}, \betal:
\batitle{Power to the people: The role of humans in interactive machine learning}.
\bjtitle{AI Magazine}
\bvolume{35}(\bissue{4}),
\bfpage{105}--\blpage{120}
(\byear{2014})
\end{barticle}
\endbibitem

\bibitem[\protect\citeauthoryear{Raji and Buolamwini}{2019}]{raji2019auditing}
\begin{bchapter}
\bauthor{\bsnm{Raji}, \binits{I.D.}},
\bauthor{\bsnm{Buolamwini}, \binits{J.}}:
\bctitle{Actionable auditing: Investigating the impact of public audits on algorithmic bias}.
In: \bbtitle{AAAI/ACM Conference on AI, Ethics, and Society (FAT)}
(\byear{2019})
\end{bchapter}
\endbibitem

\bibitem[\protect\citeauthoryear{Floridi and Cowls}{2019}]{floridi2019framework}
\begin{botherref}
\oauthor{\bsnm{Floridi}, \binits{L.}},
\oauthor{\bsnm{Cowls}, \binits{J.}}:
A unified framework of five principles for ai in society.
Harvard Data Science Review
\textbf{1}(1)
(2019)
\end{botherref}
\endbibitem

\bibitem[\protect\citeauthoryear{Bostrom}{2014}]{bostrom2014superintelligence}
\begin{bbook}
\bauthor{\bsnm{Bostrom}, \binits{N.}}:
\bbtitle{Superintelligence: Paths, Dangers, Strategies}.
\bpublisher{Oxford University Press}, \blocation{???}
(\byear{2014})
\end{bbook}
\endbibitem

\bibitem[\protect\citeauthoryear{Mittelstadt}{2019}]{mittelstadt2019principles}
\begin{barticle}
\bauthor{\bsnm{Mittelstadt}, \binits{B.D.}}:
\batitle{Principles alone cannot guarantee ethical ai}.
\bjtitle{Nature Machine Intelligence}
\bvolume{1},
\bfpage{501}--\blpage{507}
(\byear{2019})
\end{barticle}
\endbibitem

\bibitem[\protect\citeauthoryear{{European Commission}}{2021}]{eucommission2021aiact}
\begin{botherref}
\oauthor{\bsnm{{European Commission}}}:
Proposal for a Regulation Laying Down Harmonised Rules on Artificial Intelligence (AI Act).
Brussels.
COM(2021) 206 final
(2021)
\end{botherref}
\endbibitem

\bibitem[\protect\citeauthoryear{Suchman}{2007}]{suchman2007human}
\begin{bbook}
\bauthor{\bsnm{Suchman}, \binits{L.A.}}:
\bbtitle{Human-Machine Reconfigurations: Plans and Situated Actions}.
\bpublisher{Cambridge University Press}, \blocation{???}
(\byear{2007})
\end{bbook}
\endbibitem

\bibitem[\protect\citeauthoryear{Dignum}{2019}]{dignum2019responsible}
\begin{bbook}
\bauthor{\bsnm{Dignum}, \binits{V.}}:
\bbtitle{Responsible Artificial Intelligence: How to Develop and Use AI in a Responsible Way}.
\bpublisher{Springer}, \blocation{???}
(\byear{2019})
\end{bbook}
\endbibitem

\bibitem[\protect\citeauthoryear{Eubanks}{2017}]{eubanks2017automating}
\begin{bbook}
\bauthor{\bsnm{Eubanks}, \binits{V.}}:
\bbtitle{Automating Inequality: How High-Tech Tools Profile, Police, and Punish the Poor}.
\bpublisher{St. Martin’s Press}, \blocation{???}
(\byear{2017})
\end{bbook}
\endbibitem

\bibitem[\protect\citeauthoryear{Crawford}{2021}]{crawford2021atlas}
\begin{bbook}
\bauthor{\bsnm{Crawford}, \binits{K.}}:
\bbtitle{Atlas of AI: Power, Politics, and the Planetary Costs of Artificial Intelligence}.
\bpublisher{Yale University Press}, \blocation{???}
(\byear{2021})
\end{bbook}
\endbibitem

\bibitem[\protect\citeauthoryear{Christiano}{2018}]{christiano2018alignment}
\begin{botherref}
\oauthor{\bsnm{Christiano}, \binits{P.}}:
AI Alignment and Amplification.
OpenAI Blog / Alignment Forum.
\url{https://www.alignmentforum.org/posts/PaSW5asfAeaPE3xqG/ai-alignment-and-amplification}
(2018)
\end{botherref}
\endbibitem

\bibitem[\protect\citeauthoryear{Gabriel}{2020}]{gabriel2020alignment}
\begin{barticle}
\bauthor{\bsnm{Gabriel}, \binits{I.}}:
\batitle{Artificial intelligence, values and alignment}.
\bjtitle{Minds and Machines}
\bvolume{30}(\bissue{3}),
\bfpage{411}--\blpage{437}
(\byear{2020})
\end{barticle}
\endbibitem

\bibitem[\protect\citeauthoryear{Zeng et~al.}{2019}]{zeng2019linking}
\begin{botherref}
\oauthor{\bsnm{Zeng}, \binits{Y.}},
\oauthor{\bsnm{Lu}, \binits{E.}},
\oauthor{\bsnm{Huangfu}, \binits{C.}}:
Linking artificial intelligence principles.
arXiv preprint arXiv:1812.04814
(2019)
\end{botherref}
\endbibitem

\bibitem[\protect\citeauthoryear{Campbell and Gear}{1995}]{bib1}
\begin{barticle}
\bauthor{\bsnm{Campbell}, \binits{S.L.}},
\bauthor{\bsnm{Gear}, \binits{C.W.}}:
\batitle{The index of general nonlinear {D}{A}{E}{S}}.
\bjtitle{Numer. {M}ath.}
\bvolume{72}(\bissue{2}),
\bfpage{173}--\blpage{196}
(\byear{1995})
\end{barticle}
\endbibitem

\bibitem[\protect\citeauthoryear{Slifka and Whitton}{2000}]{bib2}
\begin{barticle}
\bauthor{\bsnm{Slifka}, \binits{M.K.}},
\bauthor{\bsnm{Whitton}, \binits{J.L.}}:
\batitle{Clinical implications of dysregulated cytokine production}.
\bjtitle{J. {M}ol. {M}ed.}
\bvolume{78},
\bfpage{74}--\blpage{80}
(\byear{2000})
\doiurl{10.1007/s001090000086}
\end{barticle}
\endbibitem

\bibitem[\protect\citeauthoryear{Hamburger}{1995}]{bib3}
\begin{barticle}
\bauthor{\bsnm{Hamburger}, \binits{C.}}:
\batitle{Quasimonotonicity, regularity and duality for nonlinear systems of partial differential equations}.
\bjtitle{Ann. Mat. Pura. Appl.}
\bvolume{169}(\bissue{2}),
\bfpage{321}--\blpage{354}
(\byear{1995})
\end{barticle}
\endbibitem

\bibitem[\protect\citeauthoryear{Geddes et~al.}{1992}]{bib4}
\begin{bbook}
\bauthor{\bsnm{Geddes}, \binits{K.O.}},
\bauthor{\bsnm{Czapor}, \binits{S.R.}},
\bauthor{\bsnm{Labahn}, \binits{G.}}:
\bbtitle{Algorithms for {C}omputer {A}lgebra}.
\bpublisher{Kluwer},
\blocation{Boston}
(\byear{1992})
\end{bbook}
\endbibitem

\bibitem[\protect\citeauthoryear{Broy}{1992}]{bib5}
\begin{bchapter}
\bauthor{\bsnm{Broy}, \binits{M.}}:
\bctitle{Software engineering---from auxiliary to key technologies}.
In: \beditor{\bsnm{Broy}, \binits{M.}},
\beditor{\bsnm{Denert}, \binits{E.}} (eds.)
\bbtitle{Software Pioneers},
pp. \bfpage{10}--\blpage{13}.
\bpublisher{Springer},
\blocation{New {Y}ork}
(\byear{1992})
\end{bchapter}
\endbibitem

\bibitem[\protect\citeauthoryear{Seymour}{1981}]{bib6}
\begin{bbook}
\beditor{\bsnm{Seymour}, \binits{R.S.}} (ed.):
\bbtitle{Conductive {P}olymers}.
\bpublisher{Plenum},
\blocation{New {Y}ork}
(\byear{1981})
\end{bbook}
\endbibitem

\bibitem[\protect\citeauthoryear{Smith}{1976}]{bib7}
\begin{bchapter}
\bauthor{\bsnm{Smith}, \binits{S.E.}}:
\bctitle{Neuromuscular blocking drugs in man}.
In: \beditor{\bsnm{Zaimis}, \binits{E.}} (ed.)
\bbtitle{Neuromuscular Junction. {H}andbook of Experimental Pharmacology},
vol. \bseriesno{42},
pp. \bfpage{593}--\blpage{660}.
\bpublisher{Springer},
\blocation{Heidelberg}
(\byear{1976})
\end{bchapter}
\endbibitem

\bibitem[\protect\citeauthoryear{Chung and Morris}{1978}]{bib8}
\begin{botherref}
\oauthor{\bsnm{Chung}, \binits{S.T.}},
\oauthor{\bsnm{Morris}, \binits{R.L.}}:
Isolation and characterization of plasmid deoxyribonucleic acid from Streptomyces fradiae.
Paper presented at the 3rd international symposium on the genetics of industrial microorganisms, University of {W}isconsin, {M}adison, 4--9 June 1978
(1978)
\end{botherref}
\endbibitem

\bibitem[\protect\citeauthoryear{Hao et~al.}{2014}]{bib9}
\begin{botherref}
\oauthor{\bsnm{Hao}, \binits{Z.}},
\oauthor{\bsnm{AghaKouchak}, \binits{A.}},
\oauthor{\bsnm{Nakhjiri}, \binits{N.}},
\oauthor{\bsnm{Farahmand}, \binits{A.}}:
Global integrated drought monitoring and prediction system (GIDMaPS) data sets.
figshare \url{https://doi.org/10.6084/m9.figshare.853801}
(2014)
\end{botherref}
\endbibitem

\bibitem[\protect\citeauthoryear{Babichev et~al.}{2002}]{bib10}
\begin{botherref}
\oauthor{\bsnm{Babichev}, \binits{S.A.}},
\oauthor{\bsnm{Ries}, \binits{J.}},
\oauthor{\bsnm{Lvovsky}, \binits{A.I.}}:
Quantum scissors: teleportation of single-mode optical states by means of a nonlocal single photon.
Preprint at \url{https://arxiv.org/abs/quant-ph/0208066v1}
(2002)
\end{botherref}
\endbibitem

\bibitem[\protect\citeauthoryear{Beneke et~al.}{1997}]{bib11}
\begin{barticle}
\bauthor{\bsnm{Beneke}, \binits{M.}},
\bauthor{\bsnm{Buchalla}, \binits{G.}},
\bauthor{\bsnm{Dunietz}, \binits{I.}}:
\batitle{Mixing induced {CP} asymmetries in inclusive {B} decays}.
\bjtitle{Phys. {L}ett.}
\bvolume{B393},
\bfpage{132}--\blpage{142}
(\byear{1997})
{\href{https://arxiv.org/abs/0707.3168}{{arXiv:0707.3168}}}
{[gr-gc]}
\end{barticle}
\endbibitem

\bibitem[\protect\citeauthoryear{Stahl}{2020}]{bib12}
\begin{botherref}
\oauthor{\bsnm{Stahl}, \binits{B.}}:
Deep{SIP}: Deep Learning of {S}upernova {I}a {P}arameters,
0.42,
Astrophysics {S}ource {C}ode {L}ibrary
(2020),
{\href{https://ascl.net/2006.023}{{ascl:2006.023}}}
\end{botherref}
\endbibitem

\bibitem[\protect\citeauthoryear{Abbott et~al.}{2019}]{bib13}
\begin{barticle}
\bauthor{\bsnm{Abbott}, \binits{T.M.C.}}, \betal:
\batitle{{Dark Energy Survey Year 1 Results: Constraints on Extended Cosmological Models from Galaxy Clustering and Weak Lensing}}.
\bjtitle{Phys. Rev. D}
\bvolume{99}(\bissue{12}),
\bfpage{123505}
(\byear{2019})
\doiurl{10.1103/PhysRevD.99.123505}
{\href{https://arxiv.org/abs/1810.02499}{{arXiv:1810.02499}}}
{[astro-ph.CO]}
\end{barticle}
\endbibitem

\end{thebibliography}

\end{document}